\documentclass{article}
\usepackage{spconf,amsmath,graphicx,hyperref}

\title{A Cocktail-Party Benchmark: \\Multi-Modal dataset and Comparative Evaluation Results}

\name{%
  \begin{tabular}{c}
    Thai-Binh Nguyen$^{\star}$, Katerina Zmolikova$^{\dagger}$, Pingchuan Ma$^{\dagger}$, Ngoc Quan Pham$^{\ddagger}$, \\
    \textit{Christian Fuegen}$^{\dagger}$, \textit{Alexander Waibel}$^{\star\ddagger\pm}$
  \end{tabular}%
}
\address{%
  $^{\star}$ Karlsruhe Institute of Technology, Germany \\
  $^{\ddagger}$ Carnegie Mellon University, USA \\
  $^{\pm}$ Interactive-AI LLC \\
  $^{\dagger}$ Meta AI, UK \\
  \texttt{thai-binh.nguyen@kit.edu}
}

\begin{document}

\maketitle

\begin{abstract}
We introduce the task of Multi-Modal Context-Aware Recognition (MCoRec) in the ninth CHiME Challenge, which addresses the cocktail-party problem of overlapping conversations in a single-room setting using audio, visual, and contextual cues. MCoRec captures natural multi-party conversations where the recordings focus on unscripted, casual group chats, leading to extreme speech overlap of up to 100\% and highly fragmented conversational turns. The task requires systems to answer the question ``Who speaks when, what, and with whom?'' by jointly transcribing each speaker’s speech and clustering them into their respective conversations from audio-visual recordings. Audio-only baselines exceed 100\% word error rate, whereas incorporating visual cues yields substantial 50\% improvements, highlighting the importance of multi-modality. In this manuscript, we present the motivation behind the task, outline the data collection process, and report the baseline systems developed for the MCoRec. 
\end{abstract}

\vspace{-1em}
\section{Introduction}
\label{sec:intro}
\vspace{-0.5em}

Automatic speech recognition (ASR) has achieved significant improvements in noisy and reverberant conditions over the past few decades, supported by advances in deep learning, signal processing, and the availability of large real-world corpora \cite{virtanen2012techniques, watanabe2017new, nguyen2020super, open-asr-leaderboard}. However, ASR performance in natural multi-party conversations remains limited, particularly under overlapping speech and conversational dynamics, also known as the cocktail-party setting \cite{watanabe2020chime, AliMeeting, barker18_interspeech, segbroeck20_interspeech, ami}. Although many approaches have been proposed to improve robustness through acoustic modeling, word error rates (WER) remain above 20\% in such scenarios.

In addition to research on acoustic robustness, audio-visual speech recognition (AVSR) has emerged as a promising approach by leveraging visual cues such as lip movements to complement the acoustic signal \cite{duchnowski94_icslp, suhm1999model, stiefelhagen1999modeling, bub1995knowing, yang1998visual, meier2000towards, duchnowski1995toward}. To advance this direction, several audio-visual and multi-modal datasets have been introduced. Large-scale corpora such as LRS2/LRS3 \cite{lrs2,lrs3} and Muavic \cite{muavic} focus primarily on single-speaker or Ego4D \cite{ego4d} broad egocentric scenarios, while conversational datasets such as AMI \cite{ami}, MISP \cite{misp}, and MMCSG \cite{mmcsg} capture multi-party meetings but are limited to single conversations per session. CHiME-5/6 \cite{watanabe2020chime} capture multi-party meetings but remain limited to audio-only. Despite these efforts, there remains a clear gap: audio-visual, natural, unscripted, multi-party interaction under cocktail-party conditions. The challenge in such settings arises from the so-called cocktail-party phenomenon, which describes humans’ ability to selectively attend to one conversation while ignoring others. In this context, systems must determine not only who speaks when and what is spoken, but also with whom the interaction occurs.

The MCoRec task aims to bridge the gap between previous efforts by establishing a benchmark for the cocktail-party challenge. Our dataset features audio and video recordings of up to eight participants engaged in up to four simultaneous conversations. The challenge provides a single 360° video and a single-channel audio recording, reflecting typical conditions using commercially available devices. Benchmarking systems will focus on jointly transcribing speech and clustering speakers into conversation groups.


\vspace{-1em}
\section{Dataset}
\label{sec:dataset}
\vspace{-0.5em}

The MCoRec dataset was developed externally by Interactive-AI LLC and is made available for the CHiME Challenge for research purposes. For commercial use, please contact authors. Dataset and
baseline systems can be accessed via \url{https://github.com/MCoRec/mcorec_baseline}.

\vspace{-0.5em}
\subsection{Data collection}
\label{sub:collection}
\vspace{-0.5 em}
 The recordings were conducted in natural conversational settings where participants were seated in groups around a large table. Each group remained fixed throughout a session in terms of both participants and discussion topics to ensure consistency, while allowing spontaneous and natural interactions. Within a session, groups could either sit across from each other or in separate clusters, simulating realistic social gatherings. All participants were familiar with each other to encourage fluent and unconstrained conversations. Discussion topics covered a wide range of everyday themes, including personal life, hobbies, school, work, entertainment, news, and hypothetical situations. Data collection took place across 10 different indoor environments of varying size and type, including living rooms, meeting rooms, lecture halls, and other common spaces, thereby capturing diverse acoustic and visual conditions. Each session involved up to 8 speakers and up to 4 simultaneous conversations.

\begin{figure}[htpb]
  \centering
  \includegraphics[width=0.8\linewidth]{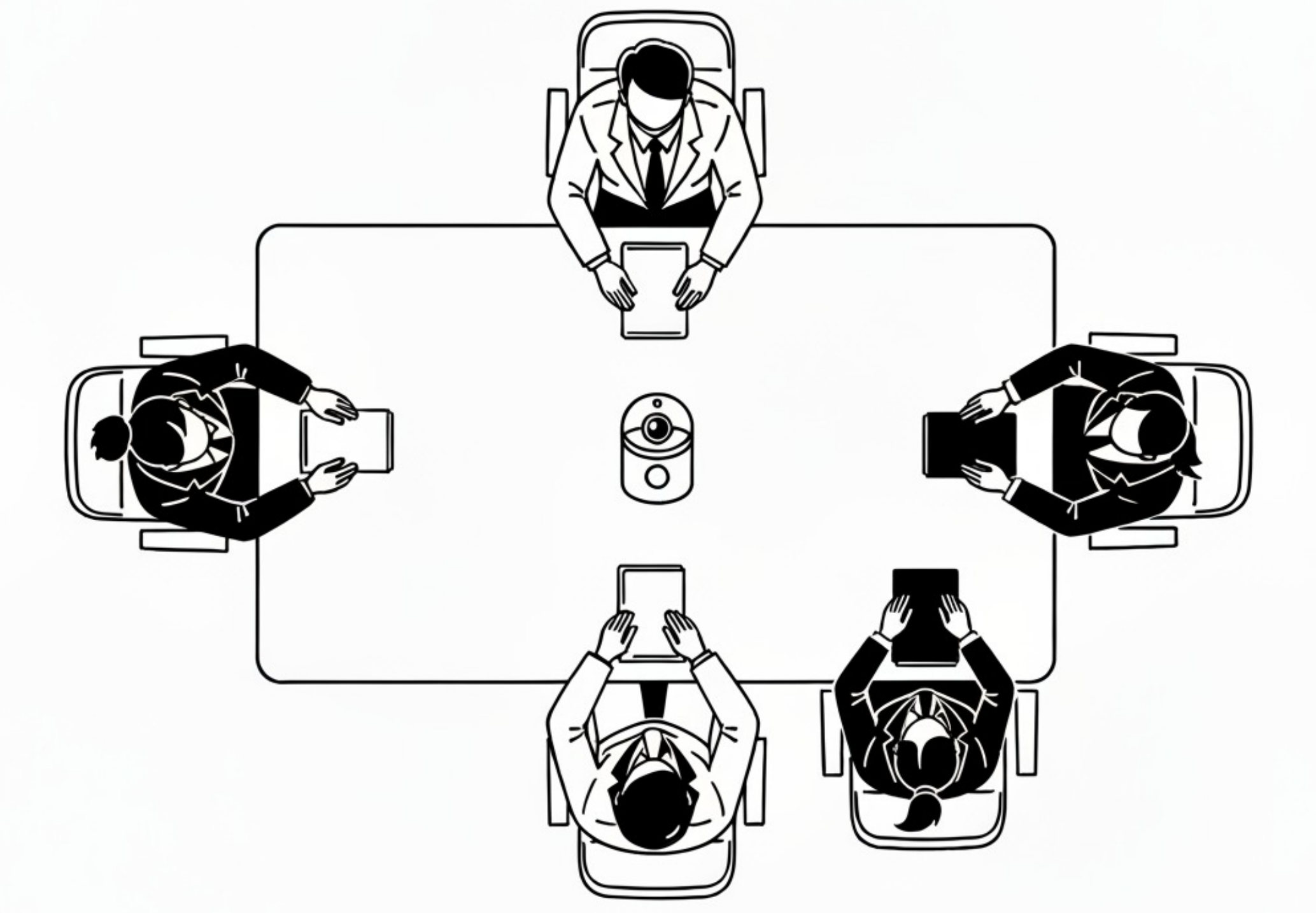}

    \caption{Recording setup of a session in the MCoRec dataset.}
  \label{fig:schema}
\vspace{-1 em}
\end{figure}

Figure \ref{fig:schema} illustrates a recording scene with two groups, one in black and the other in white, seated across from each other. Each recording session was captured using a combination of central and participant-specific devices. A 360° camera (GoPro Max) was positioned at the center of the table at seated eye level, providing a panoramic view of all participants at 4K resolution. The integrated microphone of the 360° camera was used to record the overall session audio, with distances between the camera and participants ranging from approximately 50 cm to 2 m. Simultaneously, each participant was recorded with a personal setup consisting of a smartphone camera (720p resolution) connected to a lavalier microphone (AGPTEK Z02) placed near the mouth to ensure high-quality voice capture. These cameras were positioned below the central 360° camera, below mouth level, to avoid obstructing the central view while still capturing clear visual information of the speaker’s face and mouth region. A moderator signaled the beginning and end of each session using a whistle. Each recording session lasted approximately 6 minutes.

\vspace{-0.5 em}
\subsection{Data annotation}
\label{sub:annotation}
\vspace{-0.5 em}

The annotation process was carried out in 4 main steps. In the first step, we temporally aligned all recordings using the whistle signal that marked the beginning and end of each session. Annotators manually marked the approximate region containing the whistle, after which we refined the alignment automatically. Specifically, we computed the spectral-flux onset strength envelope with the librosa library (using the $onset\_strength$ function) and identified the frame with the highest peak to pinpoint the exact whistle timestamp. This procedure ensured precise synchronization across all recording devices.

The second step focuses on transcribing the speakers' utterances. Annotators rely on the smartphone video and lapel microphone recordings, which provide the highest-quality audio, to ensure accurate transcription. Speech is manually segmented into utterances of no more than 15 seconds, each transcribed by the annotator. If a segment is unintelligible or cannot be reliably transcribed, it is marked with three asterisks (***) to indicate uncertainty.

In the third step, we performed face tracking on the 360° video to spatially locate all target participants. Since raw 360° footage is stored in a dual-fisheye format, we stitched it into a single equirectangular projection using GoPro Player. The resulting video, exported as an MP4 file with stereo audio, was used as the input for the whole recognition system. For simplicity, we continue to refer to this stitched video as the 360° video. We followed the pipeline of \cite{Chung2016OutOT}, which involves shot detection, face detection, and multi-frame tracking. This yielded a sequence of bounding boxes (top-left and bottom-right coordinates) that indicate the face of each participant throughout the recording. This step was necessary for two reasons. Firstly, not all individuals appearing in the 360° video were part of the session, so bounding boxes allowed us to restrict attention to actual participants. Secondly, face tracking enabled the correct mapping of transcriptions from Step 2 to the corresponding speaker in the video.

The fourth step was to label which speakers belonged to the same conversation. Since participants remained in fixed groups during each session, annotators assigned each speaker (based on the bounding box tracks from Step 3) a group ID. Speakers with the same group ID were considered part of the same conversation.

\vspace{-1em}
\section{Task}
\label{sec:task}
\vspace{-0.5em}

\subsection{Task Definition}
\label{sub:task_definition}
\vspace{-0.5em}

Given a 360° video $V$ and a set of target speakers 
$\mathcal{S} = \{s_1, \ldots, s_N\}$, each represented by a sequence of 
bounding boxes $\mathcal{B}_i$, the goal of the MCoRec task is to predict:
\[
f: (V, \{\mathcal{B}_i\}_{i=1}^N) \;\mapsto\; 
(\{\hat{Y}_i\}_{i=1}^N, \hat{C}),
\]
where $\hat{Y}_i$ is the transcription of speaker $s_i$ 
and $\hat{C}$ assigns speakers to conversation clusters.

\subsection{Data Partitioning}
\label{sub:datasubset}

The MCoRec dataset consists of 150 recording sessions, divided into disjoint training, development, and evaluation sets with no speaker overlap, as summarized in Table \ref{tab:dataset_stats}. The reported durations refer to the hours of 360° video. In addition, participant-specific videos recorded with smartphones amount to 26/14/32 hours for the train/dev/test splits, respectively. However, smartphone videos are only provided for the training set to support data augmentation. For the development and evaluation sets, these videos remain hidden, as the challenge focuses on recognition from the single central 360° view.

\begin{table}[h]
\centering
\begin{tabular}{ccccc}
\hline
        Dataset & sessions & conversations & duration &  speakers \\
\hline
train   & 56 & 120 & 5.6 h & 20 \\
dev     & 25 & 60  & 2.5 h & 12 \\
eval    & 69 & 157 & 6.9 h & 24 \\
\hline
\end{tabular}
\caption{Dataset statistics for train, dev, and eval splits.}
\label{tab:dataset_stats}
\end{table}
\vspace{-1.5 em}

\vspace{-1em}
\subsection{Evaluation metrics}
\label{sub:metrics}

\subsubsection{Individual speaker's WER}
\label{subsub:spk_wer}
Since the target participants are provided with bounded boxes, the transcriptions must be attributed to the correct speakers. The WER for speaker $s$ is computed as:  
\[
\text{WER}(s) = \frac{S + D + I}{W}
\]
where $S$, $D$, and $I$ are the numbers of substitutions, deletions, and insertions compared to the reference transcript, and $W$ is the total number of words in the reference. The final score is the average WER across all speakers.

\vspace{-1em}
\subsubsection{ Conversation Clustering Performance}
\label{subsub:cc_per}
The system outputs a cluster assignment for each session. For each unordered pair of speakers $\{i,j\}$, a true positive (TP) occurs if they are correctly assigned to the same conversation, a false positive (FP) if they are assigned to the same conversation but should not be, and a false negative (FN) if they should be in the same conversation but are not. Precision and recall are defined as 
\[
\text{Precision} = \frac{TP}{TP + FP}, \quad \text{Recall} = \frac{TP}{TP + FN}
\]

The pairwise $F1$ score is:
\[
F1 = \frac{2 \cdot \text{Precision} \cdot \text{Recall}}{\text{Precision} + \text{Recall}}
\]
These are averaged across all sessions.

\subsubsection{Joint ASR-Clustering Error Rate}
Our primary metric combines ASR and clustering into a single score. This is defined to provide a final ranking of participants in the challenge.  
For each speaker $s$, we compute
\[
\text{JointError}(s) = 0.5\,\text{WER}(s) + 0.5\,(1 - F1(s))
\]
where $\text{WER}(s)$ is the speaker-dependent word error rate defined in section \ref{subsub:spk_wer} and $F1(s)$ measures clustering quality specifically for speaker $s$.  

To compute $F1(s)$, we adopt a one-vs-rest approach: we consider all unordered pairs that include $s$, and compare the predicted clustering against the ground truth. A TP occurs when $s$ and another speaker are correctly assigned to the same conversation, while FP and FN capture incorrect or missing assignments. Precision, recall, and $F1(s)$ are then computed based on these pairs same as in section \ref{subsub:cc_per}.  

This metric ranges from 0 (perfect) to 1 (worst), providing a balanced evaluation of both transcription accuracy and clustering quality. The final score is the average of $\text{JointError}(s)$ across all speakers in all sessions.



\vspace{-0.5em}
\section{Baselines}
\label{sec:baselines}
\vspace{-0.5em}

The baseline is a cascade system with three main components. Active Speaker Detection determines when a speaker is talking. Audio-Visual Speech Recognition generates the transcription. Conversation Clustering predicts who is speaking with whom.





\vspace{-1em}
\subsection{Active Speaker Detection (ASD)} \label{sub:asd} 

Each recording session lasts about 6 minutes, and in a conversation, speakers do not talk continuously. It is therefore necessary to detect the segments where a speaker is active, so that recognition is applied only to these parts. This reduces computation and improves performance, as shown in the AVCocktail study \cite{nguyen25b_interspeech}. For the baseline, we reuse the ASD model \cite{ASD}, which was also effective in the AVCocktail study. This model uses a lightweight CNN-based to extract audio and visual features then fused and processed by a GRU-based detector to predict active speaking at the frame level.

On the development set, the model achieves a Micro-Averaged Intersection over Union (IoU) of 75.58\%. IoU measures how much the predicted active segments overlap with the annotated ground-truth segments. Although this metric is not used for benchmarking systems in the challenge, it provides useful insight into the performance of the ASD module as an intermediate step. A value of 75.58\% indicates that the model can localize active speaking regions with reasonable accuracy, but also leaves room for improvement, especially in challenging multi-speaker scenarios.

\vspace{-1em}
\subsection{Audio-Visual Speech Recognition (AVSR)}
\label{sub:avsr}
\vspace{-0.5 em}



Candidate segments from the Active Speaker Detection module are then processed by an Audio-Visual Speech Recognition (AVSR) system. To evaluate our MCoRec dataset, we compare several recent baselines. These include AV-HuBERT CTC/Attention \cite{nguyen25b_interspeech}, which uses a pretrained AV-HuBERT encoder with a CTC/Attention decoder; AutoAVSR \cite{autoavsr}, which uses a Conformer encoder to process both audio and video, and a CTC/Attention decoder; Muavic-EN \cite{anwar23_interspeech}, which combines a pretrained AV-HuBERT encoder with a Transformer decoder; and Llama-AVSR \cite{llamaavsr}, which uses AV-HuBERT for encoding visual input and Whisper-medium for encoding audio input, and feeds both into Llama-3.1-8B for transcription.

In addition to using off-the-shelf models from previous studies, we also fine-tuned an AV-HuBERT model with a CTC/Attention decoder on the MCoRec training set. The training data in MCoRec consists of synchronized 360° videos and participant-specific videos, with alignment available between the two streams. To leverage this setup, we applied a simple augmentation strategy: for each segment, we paired the audio and visual streams from both the 360° view and the corresponding participant-specific view that shared the same start and end times. This pairing effectively doubles the usable training material and helps the model generalize better across different viewpoints. In total, this process yielded around 104 hours of audio-visual data for fine-tuning.

\begin{table}[h]
\centering
\begin{tabular}{lc}
\hline
Model & WER \\
\hline
AV-HuBERT CTC/Attention & 55.36 \\
Muavic-EN               & 71.80 \\
Auto-AVSR               & 83.15 \\
Llama-AVSR               & 77.60 \\
AV-HuBERT CTC/Attention(MCoRec finetuned) & 49.90 \\
\hline
\end{tabular}
\caption{Speaker's WER for different AVSR models on the MCoRec development set.}
\label{tab:wer_results}
\end{table}

Table \ref{tab:wer_results} shows the results of different AVSR models on the MCoRec development set. Among the off-the-shelf baselines, AV-HuBERT CTC/Attention performs best with a WER of 55.36\%. When comparing the pretrained AV-HuBERT CTC/Attention model against our fine-tuned version on the MCoRec training set, we observe a clear performance improvement. The word error rate (WER) dropped from 55.36\% without fine-tuning to 49.90\% after fine-tuning, representing an absolute reduction of 5.46\% and a relative improvement of approximately 9.9\%. This gain highlights the benefit of adapting the model to the MCoRec domain, where multi-party conversational speech is captured with both 360° and participant-specific views. An analysis of the errors reveals that insertion errors account for the majority, indicating that the model tends to over-generate words in complex multi-speaker conditions. Despite the improvements from fine-tuning, the relatively high WERs indicate that there is still substantial room for improvement for AVSR systems under realistic conversational conditions. 

\vspace{-1em}
\subsection{Conversation Clustering}
\label{sub:conversation_clustering}
\vspace{-0.5 em}

The underlying assumption is that speakers in the same conversation usually take turns, while speakers in different conversations often talk at the same time. Based on this assumption, we design the baseline conversation clustering module using a time-based approach that analyzes temporal speaking patterns and overlaps. The input to this module is the active speaker detection output (detail in section \ref{sub:asd}), which provides time segments for each active speaker.

First, speaker activity is extracted from the ASD outputs to determine when each speaker is talking. Based on these activity segments, we calculate pairwise conversation scores between speakers. The score reflects how likely two speakers belong to the same conversation. Specifically, temporal overlap between two speakers is treated as evidence that they are in different conversations, while sequential, non-overlapping speech is treated as evidence that they are in the same conversation. The score is defined as:  

\vspace{-0.5 em}
\begin{equation}
\text{Score}(i,j) = 1 - \frac{\text{overlap\_duration}(i,j)}{\text{total\_duration}(i,j)}
\end{equation}
\vspace{-0.5 em}

where $\text{overlap\_duration}(i,j)$ is the total time both speakers $i$ and $j$ talk simultaneously, and $\text{total\_duration}(i,j)$ is the union of their speaking times.  

These pairwise scores are then converted into a distance matrix, where higher scores (less overlap) correspond to smaller distances, indicating a higher likelihood of being in the same conversation. Finally, agglomerative clustering is applied on the distance matrix to group speakers into conversation clusters. The distance\_threshold for agglomerative clustering is 0.3. Which means the algorithm will only merge speakers into the same cluster if their speech overlaps for less than 30\% of their combined speaking time. In the development set, the conversation clustering's $F1$ Score is 0.8153. Detail of how the $F1$ score been calculated is in section \ref{subsub:cc_per}.


\vspace{-1em}
\section{Conclusion}
\label{sec:conclusion}
\vspace{-0.5 em}

The ``MCoRec'' challenge is designed to advance research in multi-modal, context-aware speech recognition under realistic multi-talker conditions. This paper has introduced the first edition, which focuses on conversational speech in cocktail-party scenarios. Our best baseline system achieved a Joint ASR-Clustering Error Rate of 0.3548, indicating that there is still substantial room for improvement. The dataset and baseline systems have been made publicly available to foster reproducibility and collaboration. Submitted systems and results will be presented at the CHiME-9 workshop as a satellite event of ICASSP 2026.
\vspace{-1em}
\section{Acknowledgment}
\vspace{-0.5em}
The authors gratefully acknowledge support from the EU’s Horizon research \& innovation programme (101135798 – Meetween; 101213369 – DVPS). We thank Interactive-AI LLC for providing the database for research purposes.
\vspace{-1em}
\bibliographystyle{IEEEbib}
\bibliography{strings,refs}

\end{document}